\documentclass[runningheads]{llncs}
\usepackage[T1]{fontenc}
\usepackage{graphicx}
\usepackage{booktabs}
\usepackage{amsmath}
\usepackage{amssymb}
\usepackage[misc]{ifsym}
\newcommand{\corr}{(\Letter)}
\usepackage{mwe}
\usepackage{float}
\usepackage{adjustbox}
\usepackage{multirow}
\usepackage{subcaption}
\usepackage{hyperref}

\begin{document}

\title{Saliency-Guided Replay for Continual Learning in Deep Spiking Neural Networks}

\titlerunning{Saliency-Guided Replay for Continual Learning in Deep SNNs}

\author{Francesco Panetta\inst{1} \and
Federico Di Valerio\inst{1} \corr \and
Michela Proietti\inst{2}}

\authorrunning{F. Panetta et al.}

\institute{Sapienza University of Rome, Italy \email{panetta.1950468@studenti.uniroma1.it, divalerio@diag.uniroma1.it}
\and
Goethe University Frankfurt, Germany \email{m.proietti@em.uni-frankfurt.de}}

\maketitle              

\begin{abstract}
Spiking Neural Networks (SNNs) are promising for energy-efficient neuromorphic computing, but they still suffer from catastrophic forgetting when learning tasks sequentially. This work investigates the integration of Experience Packing and Replay (EPR) as a memory-efficient continual learning strategy for deep SNNs under low-latency constraints. The proposed framework combines  Batch Normalization Through Time-stabilized VGG9 SNNs with Spike Activation Maps (SAM), which provide forward-only explanations based on spike activity. These explanations are temporally aggregated into spatial attention maps and used to select the most informative image patches for replay, instead of storing full images. Experiments on Split CIFAR10 and Split CIFAR100 show that SAM-guided EPR improves the accuracy–memory trade-off under tight memory budgets, and remains competitive while requiring substantially less storage. The code is available at \href{https://github.com/KRLGroup/SGR-SNN}{this link}.

\end{abstract}

\section{Introduction}
Artificial intelligence systems deployed in real-world environments must adapt to evolving data while meeting strict computational and energy constraints. Yet standard neural networks still suffer from catastrophic forgetting, where learning new tasks disrupts previously acquired representations \cite{mccloskey1989catastrophic}. Continual learning (CL) addresses this by preserving prior knowledge while integrating new information over time. Among CL strategies, replay-based methods are particularly effective, storing representative samples from past tasks and mixing them with current data during training \cite{buzzega2020dark,rolnick2019experience}. However, their performance depends on memory capacity: larger buffers improve retention but are impractical for resource-constrained systems. To reduce this cost, Experience Packing and Replay (EPR) stores saliency-guided image patches rather than full samples, substantially reducing memory requirements while maintaining competitive CL performance \cite{saha2023saliency}.

Spiking Neural Networks (SNNs) have emerged as a promising paradigm for energy-efficient machine intelligence, using sparse spike events rather than continuous-valued activations \cite{maass1997networks}. Surrogate-gradient learning has enabled deep SNNs to achieve competitive performance on vision benchmarks \cite{neftci2019surrogate,TAVANAEI201947}, while Batch Normalization Through Time (BNTT) improves training stability and reduces inference latency \cite{kim2021revisiting}. However, continual learning in SNNs remains underexplored \cite{minhas2025continual}. Replay-based approaches have only recently been adapted to SNNs, first through standard experience replay to mitigate catastrophic forgetting \cite{proiettim_cl}, and more recently through compact latent replay to reduce storage and computation costs in neuromorphic systems \cite{dequino2024compressed}. To the best of our knowledge, no prior work has used explainable artificial intelligence (XAI) to guide SNN learning and mitigate catastrophic forgetting, despite its effectiveness in artificial neural networks (ANNs) \cite{proietti2025xai}.

In this work, we investigate whether temporally grounded saliency signals in SNNs can guide memory-efficient replay. Specifically, we adapt EPR \cite{saha2023saliency} to deep SNNs by replacing conventional ANN saliency mechanisms with Spike Activation Maps (SAM) \cite{kim2021visual}, which produce forward-only explanations from inter-spike interval dynamics. Using BNTT-trained VGG9 SNNs \cite{kim2021revisiting} operating at low latency (25 timesteps) on Split CIFAR10 and Split CIFAR100 CL benchmarks, we show that saliency-guided replay can be effectively transferred to the spiking domain, improving performance under equal memory budgets and reducing forgetting in low-memory regimes.

\section{Related Work}
SNNs provide a sparse, event-driven alternative to conventional ANNs, making them particularly attractive for neuromorphic and energy-efficient learning systems. Modern SNNs can be trained either through ANN-to-SNN conversion or directly with surrogate-gradient methods, enabling competitive performance across a range of vision and learning tasks~\cite{diehl2015fast,neftci2019surrogate,sengupta2019going,TAVANAEI201947}. Continual learning in SNNs has recently attracted increasing attention, motivated by the need for adaptive neuromorphic systems that can learn over time under memory and energy constraints~\cite{mishra2023survey}. Existing approaches mostly focus on regularization-based mechanisms to mitigate catastrophic forgetting, including adaptive synaptic plasticity, Bayesian learning, and sparse selective activation~\cite{Panda_2018,Shen_Ni_Xu_Tang_2024,skatchkovsky2022bayesian}. In contrast, replay-based continual learning for SNNs remains under-explored~\cite{proiettim_cl}. Recent work on compressed latent replay for SNNs further confirms the relevance of rehearsal-based strategies in spiking models, while highlighting the need for memory-efficient replay mechanisms tailored to temporal spiking representations~\cite{dequino2024compressed}. In parallel, recent continual learning methods for conventional ANNs have begun to exploit XAI to identify and preserve task-relevant information from previous experiences~\cite{bai2023saliency,di2026cip,ebrahimi2021remembering,rymarczyk2023icicleinterpretableclassincremental,saha2023saliency}. Among these, EPR~\cite{saha2023saliency} is particularly relevant to our work: it uses saliency maps to select discriminative image patches and stores only those patches in the replay buffer. This reduces memory requirements while maintaining competitive continual learning performance. However, most saliency-guided replay methods rely on explanation mechanisms originally designed for conventional ANNs. Applying these mechanisms directly to SNNs overlooks the distinctive properties of spiking computation, including sparse event-driven activity and temporal dynamics. This motivates spike-compatible explanation strategies that can exploit temporal spiking activity~\cite{bitar2023gradient,kim2021visual}. Our work addresses this gap by adapting the principle of saliency-guided replay to SNNs. Instead of using Grad-CAM-based image patches as in EPR, we use SAM-based spatio-temporal saliency \cite{kim2021visual} to select compact replay information. In this way, explanation becomes a functional component of continual learning in SNNs, rather than only a diagnostic tool.

\section{Methods}
This section introduces a continual learning framework for SNNs that combines Spike Activation Maps (SAM) \cite{kim2021visual} with a replay strategy inspired by Experience Packing and Replay (EPR) \cite{saha2023saliency}. The key idea is to use temporally grounded saliency to store only the most informative regions of past samples, enabling memory-efficient replay in neuromorphic continual learning.

\subsection{Spike Activation Maps}
\label{sec:sam}
Standard gradient-based explanation methods, such as Grad-CAM \cite{selvaraju2017grad}, are difficult to apply to SNNs because of their discrete and non-differentiable spike dynamics. We therefore adopt SAM, a forward-only explanation mechanism that extracts saliency directly from spike-train activity.\\

Given a spike tensor $S \in \{0,1\}^{T \times C \times H \times W}$, where $T$ is the number of timesteps, $C$ the number of channels, and $H \times W$ the spatial resolution, SAM is computed using a causal temporal kernel:
\begin{equation}
K(\tau) =
\begin{cases}
\exp(-\gamma \tau), & \tau > 0,\\
0, & \tau = 0,
\end{cases}
\end{equation}

where $\gamma$ controls the decay rate. This kernel assigns larger weights to recent spikes while ensuring that only past activity contributes to the current timestep. Next, a neuronal contribution score is computed:

\begin{equation}
\mathrm{NCS}_{c,h,w}(t)
=
\sum_{\tau=1}^{t-1}
K(\tau)\,S_{c,h,w}(t-\tau).
\end{equation}

Finally, the SAM heatmap at timestep $t$ is obtained by combining the accumulated spike history with the current spike activity:

\begin{equation}
H_t(h,w)
=
\sum_{c=1}^{C}
\mathrm{NCS}_{c,h,w}(t)\,
S_{c,h,w}(t).
\end{equation}

The resulting tensor $H \in \mathbb{R}^{T \times H \times W}$ captures the spatio-temporal importance of neural activations during the forward pass. For replay selection, we reduce the temporal dimension either by averaging all heatmaps, $H=\frac{1}{T}\sum_{t=1}^{T} H_t$, or by using only the final heatmap, $H=H_T$. Finally, SAMs can be extracted from different network layers. Based on findings on deep convolutional networks \cite{zeiler2014visualizing}, we use the deepest convolutional layer, which provides lower spatial resolution but richer semantic representations and more discriminative saliency maps.

\subsection{Experience Packing and Replay}
We propose an SNN-adapted EPR strategy that uses SAMs to identify and store the most salient image patch instead of the full input. Each buffer exemplar contains the extracted patch, its original coordinates $(top, left)$, the class label, the aggregated SAM heatmap used for selection and optionally a SAM history for offline explanation-drift analysis. 
For each input, we compute an aggregated SAM heatmap
$\mathrm{H} \in \mathbb{R}^{H_{sam} \times W_{sam}}$ at the target layer and bilinearly upsample it to the input resolution, yielding
$\mathrm{H}_{full} \in \mathbb{R}^{32 \times 32}$ for CIFAR images. We then scan the image with a sliding window of size $P \times P$, with $P=24$ by default, and compute the saliency score of each candidate patch:

\begin{equation}
s_{(top,left)} = \sum^{top+P}_{h=top} \sum^{left+P}_{w=left} \mathrm{H}_{full} (h,w). 
\end{equation}

The most informative patch is selected by choosing the candidate location with the maximum saliency score:

\begin{equation}
(top^*,left^*) = arg \space \max_{(top,left)} s_{(top,left)}
\end{equation}

The selected patch is then extracted from the original image as:

\begin{equation}
patch = image[:, \space top^* : top^*+ P, \space left^* : left^*+P].
\end{equation}

During replay, the patch is placed back into a zero-valued canvas of the original input size at its stored coordinates $(top^*,left^*)$.
After each task, the memory buffer is updated using a temporary ring buffer $\mathrm{M}_t$ of recent samples from the current task. For Task $1$, the model computes the SAM heatmap, extracts the most salient patch, reconstructs its padded version, and stores the exemplar only if the prediction is correct or if the ground-truth class appears among the top-$3$ predictions. For later tasks, the same procedure is applied to new samples, while stored exemplars may optionally be reprocessed with the updated model to track explanation drift. We refer to this EPR-inspired variant \cite{saha2023saliency} as EPR-SP. As an ablation, we also consider BASE-EPR, where the samples in $\mathrm{M}_t$ are drawn uniformly at random from the current task training set.

At training stage $i$, the model is optimized using current-task mini-batches $B_i \subset D_i$ and replay minibatches $B_{mem}$ sampled from the previously stored patches. Since replay samples are reconstructed as padded inputs $\tilde{x}$, the same supervised objective can be applied to both current and replay data:

\begin{equation}
L_{total} = L^{(i)}_{CE} + \lambda_{replay}L_{replay},
\end{equation}
where
\begin{equation}
L^{(i)}_{CE} = \frac{1}{|B_i|} \sum_{(x,y) \in B_i} \ell_{CE}(f_{\theta}(x),y), \quad \quad L_{replay} = \frac{1}{|B_{mem}|} \sum_{(\tilde{x},y) \in B_{mem}} \ell_{CE}(f_{\theta}(\tilde{x}),y).
\end{equation}

Here $\lambda_{replay}$ controls the stability-plasticity trade-off between retaining knowledge from previous tasks and adapting to the current one. This formulation avoids an explicit explanation-matching loss and instead uses explanations only to guide exemplar selection, making the replay mechanism compatible with the class-agnostic nature of SAM.

\subsection{Evaluation Metrics}
\label{sec:metrics}

Let $a_{i,j}$ denote the accuracy on the test set of task~$j$ after the network has completed training on task~$i$, and let $N$ be the total number of tasks. We evaluate the models' final performance using the average accuracy and forgetting:
\begin{equation}
\text{AvgAcc} = \frac{1}{N}\sum_{j=1}^{N} a_{N,j}, \qquad \text{AvgForg} = \frac{1}{N-1}\sum_{j=1}^{N-1} \bigl(a_{j,j} - a_{N,j}\bigr)
\end{equation}

\section{Experimental Setup}
\label{sec:experimental-setup}

We perform experiments on Split-CIFAR10 (5 tasks) and Split-CIFAR100 (both the 10-task and the 20-task versions) \cite{cifar}. For each task, $20\%$ of the task training data is held out as a validation set used for early stopping. The used architecture is taken from \cite{kim2021revisiting} and consists of a VGG-9-inspired SNN. Convolutional layers are made of leaky integrate-and-fire neurons \cite{lif1} with a fixed threshold of $1.0$ and soft-reset dynamics, and are paired with per-timestep batch normalization \cite{kim2021revisiting}.
Input images are encoded as Poisson spike trains over $T{=}25$ discrete timesteps. A two-layer classification head outputs task-specific logits via independent task-specific linear layers, each mapping the shared feature representation to the classes of its corresponding task. We consider both task- (TIL) and class-incremental (CIL) learning scenarios, where task identities are available during both training and testing and just during training, respectively. In the CIL scenario, all task heads are concatenated at inference time and the class with the highest logit across all seen tasks is selected. 
Since the heads from previous tasks are frozen while training on the current task, we add a separate replay phase after each training epoch, in which the heads are unfrozen and samples from previous tasks are replayed.
We compare EPR variants, BASE-EPR and EPR-SP, with standard experience replay (identified as BASELINE), which stores full $32\times 32$ images sampled randomly from the current task's training set.

\subsection{Hyperparameter Search}

\begin{figure}[t]
    \centering
    \includegraphics[width=0.9\linewidth]{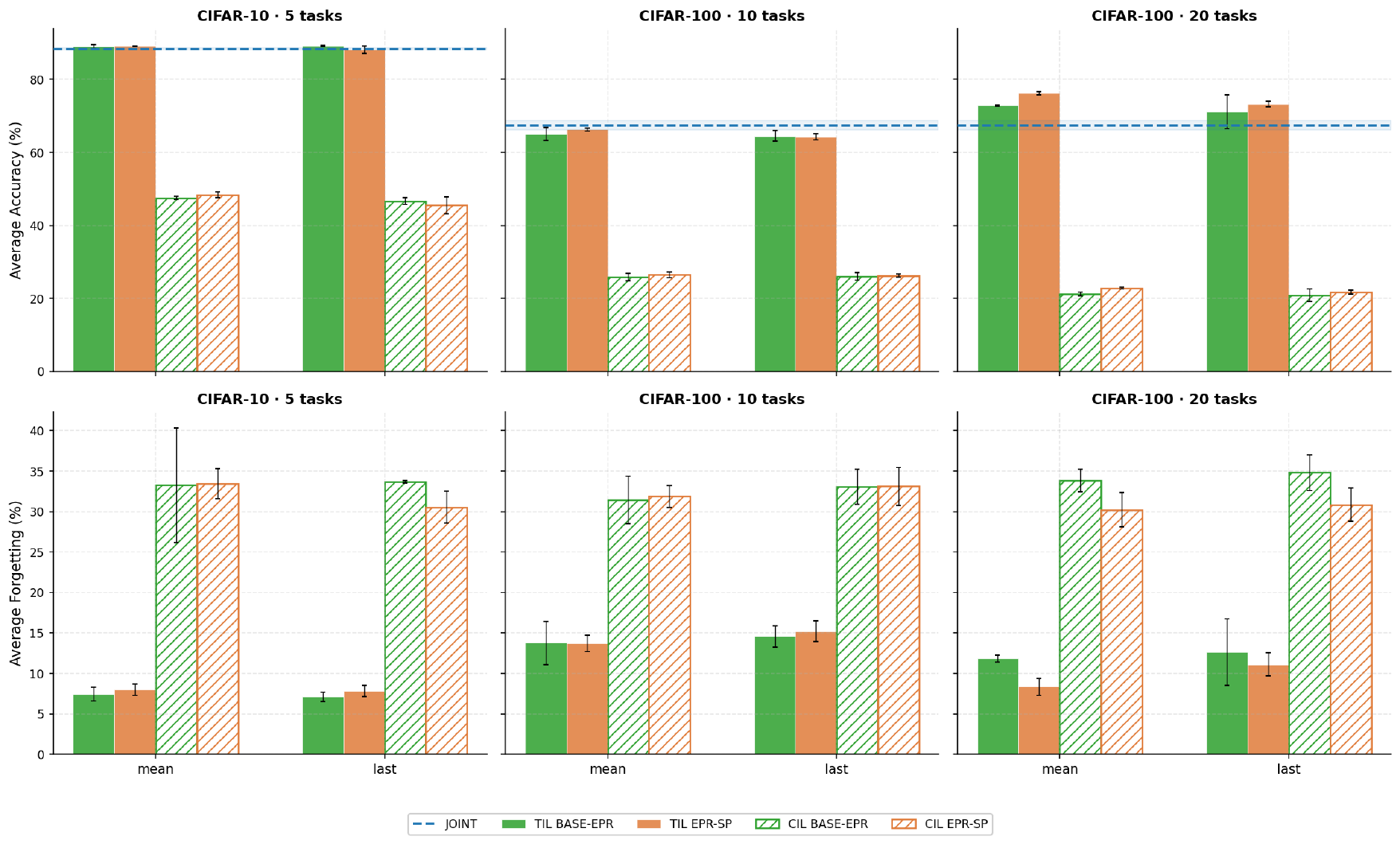}
    \caption{Comparison of BASE-EPR and EPR-SP using mean and last temporal aggregation across benchmarks. Columns denote benchmark settings, with \textit{AvgAcc} (top) and \textit{AvgForg} (bottom). Solid and hatched bars indicate TIL and CIL, respectively; error bars show standard deviation over 3 seeds.
    }
    \label{fig:grid}
\end{figure}
All models are trained using mini-batch SGD \cite{Robbins1951} with momentum $0.9$, batch size $128$, weight decay $5{\times}10^{-4}$, gradient clipping $\ell_2{\leq}1.0$, and initial learning rate $\eta \in \{0.1, 0.3\}$. These parameters were selected by performing a grid search on the BASELINE, varying membrane leak $\lambda_{\text{mem}} \in \{0.95,0.99\}$, learning rate $\eta \in \{0.1, 0.3\}$, and dropout $\in \{\text{on, off}\}$ at a fixed memory budget ($M_t=100$ samples for CIFAR10, $M_t=30$ samples for CIFAR100). Subsequently, we run an ablation to choose sample selection strategy (BASE-EPR vs EPR-SP) and SAM aggregation method (mean vs last, as described in Section \ref{sec:sam}). Figure \ref{fig:grid} shows that across all three benchmarks EPR-SP consistently outperforms or ties its BASE-EPR counterpart, confirming that model-validated selection improves the discriminative quality of the replay buffer. Regarding temporal aggregation, \emph{mean} yields consistently higher or comparable accuracy across all benchmarks. Finally, replay-weight ablations ($\lambda_r=0.3,0.5,0.7,1.0$) showed that $\lambda_r=1.0$ performs best on CIFAR10, whereas $\lambda_r=0.3$ provides a better stability–plasticity trade-off on CIFAR100. These values are used in subsequent experiments. Therefore, in all subsequent experiments, we use EPR-SP with mean aggregation as the default configuration, with $\lambda_r=1.0$ for CIFAR10 (in all scenarios) and CIFAR100 with tight memory budgets ($M_t=30$), and $\lambda_r=0.3$ for CIFAR100 with larger memory budgets ($M_t=53$). The learning rate follows a cosine annealing schedule within each task, decaying from $\eta$ to a minimum of $10^{-4}$. At each task boundary, the schedule and the learning rate are reset to the initial value $\eta$. Early stopping monitors validation accuracy on the current task with patience $10$ for CIFAR10 over $50$ epochs and $20$ for CIFAR100 over $100$ epochs. Finally, concerning heatmap computation, we compute SAMs from layer~$7$ with temporal decay $\gamma{=}0.7$, which experimentally gave the best results.

\begin{figure}[t]
  \centering
  \includegraphics[width=0.95\linewidth]{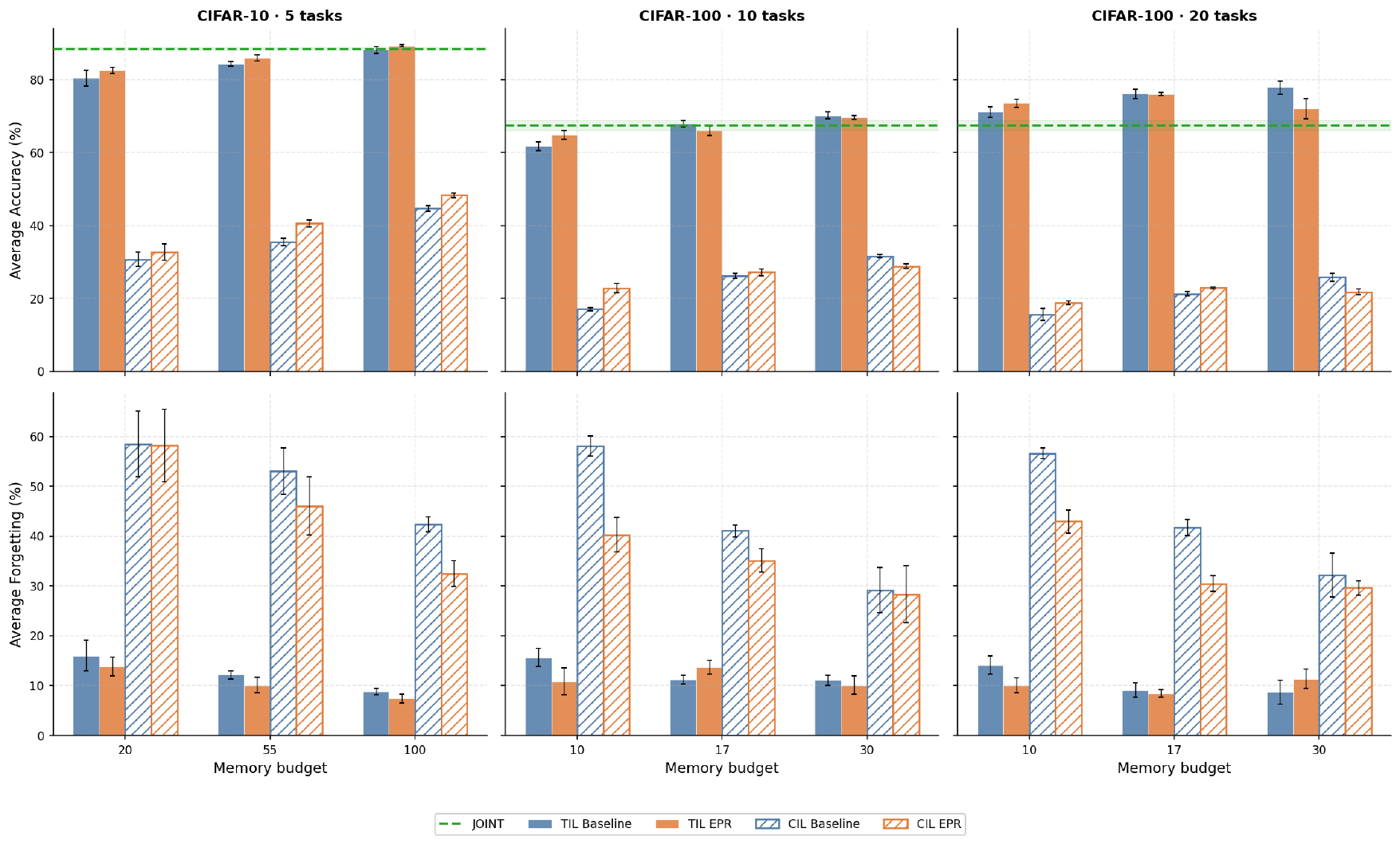}
  \caption{%
    TIL and CIL accuracy versus memory budget for BASELINE and EPR-SP across all benchmarks. At equal budgets, EPR-SP stores more patches than BASELINE stores full samples, yielding higher accuracy at tight and moderate budgets, with the advantage diminishing as memory increases.
  }
  \label{fig:acc-vs-eff-mem}
\end{figure}

\section{Results}
\label{sec:results}

\subsection{Comparison Under Equal Memory Budgets}
\label{sec:res-budget}

A $24{\times}24$ patch requires $3{\times}24{\times}24{=}1{,}728$ bytes versus $3{\times}32{\times}32{=}3{,}072$ for a full image, yielding a storage cost factor of $\approx 0.56$ ($\approx44\%$ reduction).
To enable fair comparisons between the BASELINE and EPR-SP we define:
\begin{equation}
  \text{Memory budget $(M_B)$} = M_t \times \text{cost}, \qquad
  \text{cost} =
  \begin{cases}
    1.00 & \text{BASELINE} \\
    0.56 & \text{BASE-EPR, EPR-SP}
  \end{cases}
  \label{eq:eff_mem}
\end{equation}
For instance, EPR-SP at $M_t=177$ ($M_B$=$99$) is directly comparable to the BASELINE at $M_t=100$ ($M_B$=$100$). Figure~\ref{fig:acc-vs-eff-mem} shows the results under comparable memory budgets. EPR-SP outperforms the BASELINE at the tightest memory budget across all datasets. On CIFAR100/10 and CIFAR100/20 (10 and 20 tasks respectively) at $\text{$M_B$}\approx10$, the \textit{AvgAcc} increases by $3.1$ and $2.4$ points in TIL and $5.8$ and $3.2$ points in CIL, with CIL \textit{AvgForg} reduced by $17.8$ and $13.7$ points. The gap narrows as the budget grows: at $\text{$M_B$}\approx30$ on CIFAR100, the BASELINE recovers the lead on TIL accuracy while EPR-SP retains a forgetting advantage, suggesting the occurrence of a buffer-saturation effect in which additional low-resolution samples yield diminishing
returns compared to full-resolution ones.
These results confirm that in the presence of stringent memory constraints, as in the case of neuromorphic hardware, SAM-guided patch replay can obtain better results than standard replay.

\subsection{Comparison Under Equal Sample Budgets}
\begin{figure}[t]
  \centering
  \includegraphics[width=0.95\linewidth]{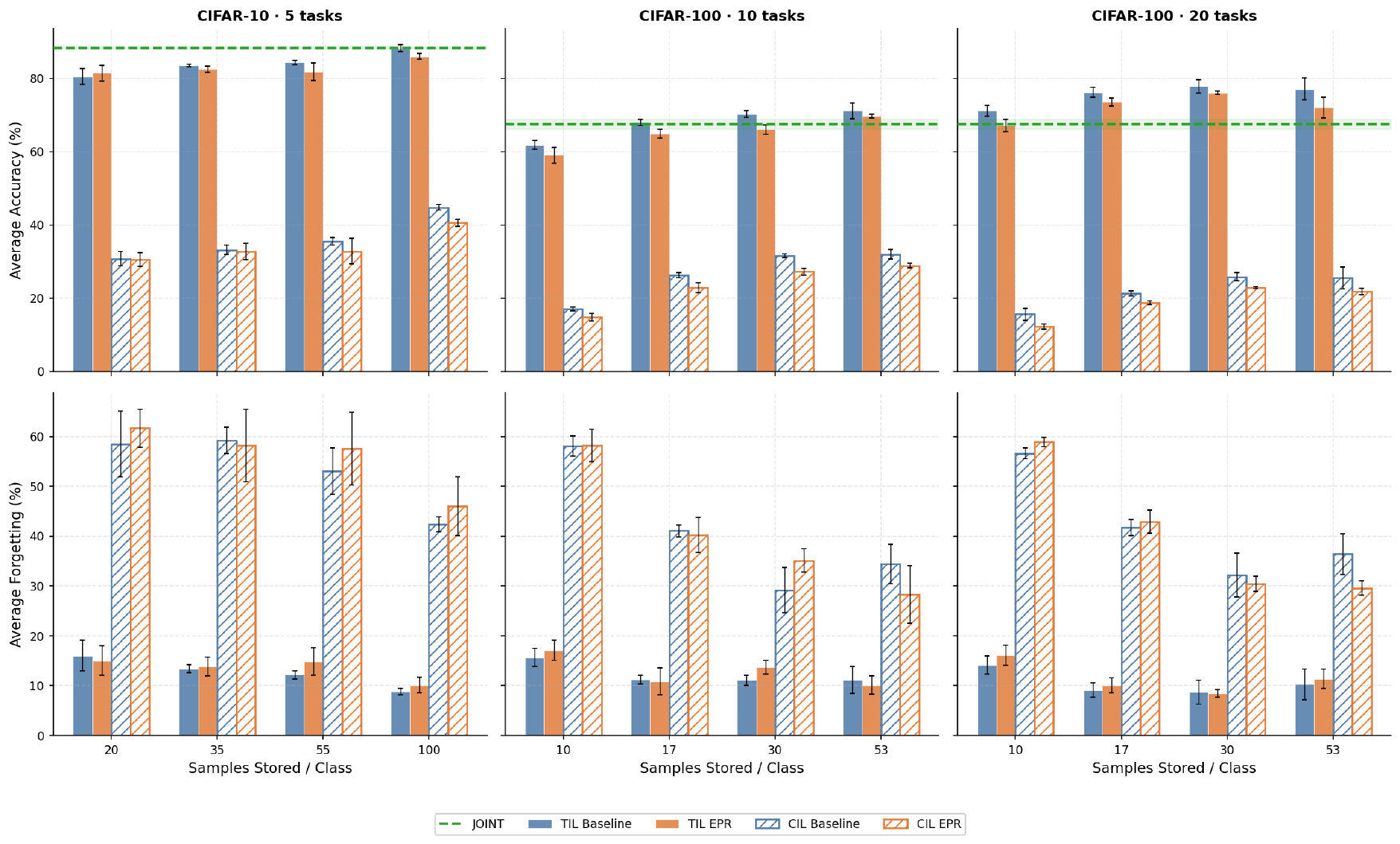}
  \caption{%
    TIL and CIL accuracy versus exemplars per class $K$ for BASELINE and EPR-SP on Split-CIFAR10 (5 tasks), CIFAR100/10 (10 tasks), and CIFAR100/20 (20 tasks). At the same $K$, EPR-SP uses $44\%$ less memory than BASELINE. The gap narrows as $K$ decreases, with EPR-SP outperforming BASELINE at the tightest CIFAR10 budget.
  }
  \label{fig:acc-vs-samples}
\end{figure}

Figure~\ref{fig:acc-vs-samples} further highlights the practical benefit of patch-based replay when the memory budget is extremely limited. When performance is plotted against the number of stored samples per class, the BASELINE generally achieves higher accuracy than EPR-SP at the same nominal number of samples, especially as the buffer size increases. However, an EPR-SP sample requires only about $56\%$ of the memory of a full image, therefore the resulting performance gap is relatively small compared to the memory savings. Overall, EPR-SP preserves competitive TIL and CIL accuracy while reducing the storage cost by $44\%$. This suggests that, in highly memory-constrained settings, EPR-SP offers a favorable trade-off: it sacrifices only a limited amount of accuracy while enabling a much more compact replay buffer. This behavior is particularly relevant for neuromorphic and embedded scenarios, where memory is often a stronger constraint than raw computational capacity.

\subsection{Explanation Drift Analysis}

\begin{figure}[t]
    \centering
     \begin{subfigure}{0.48\linewidth}
         \centering
         \includegraphics[width=\linewidth]{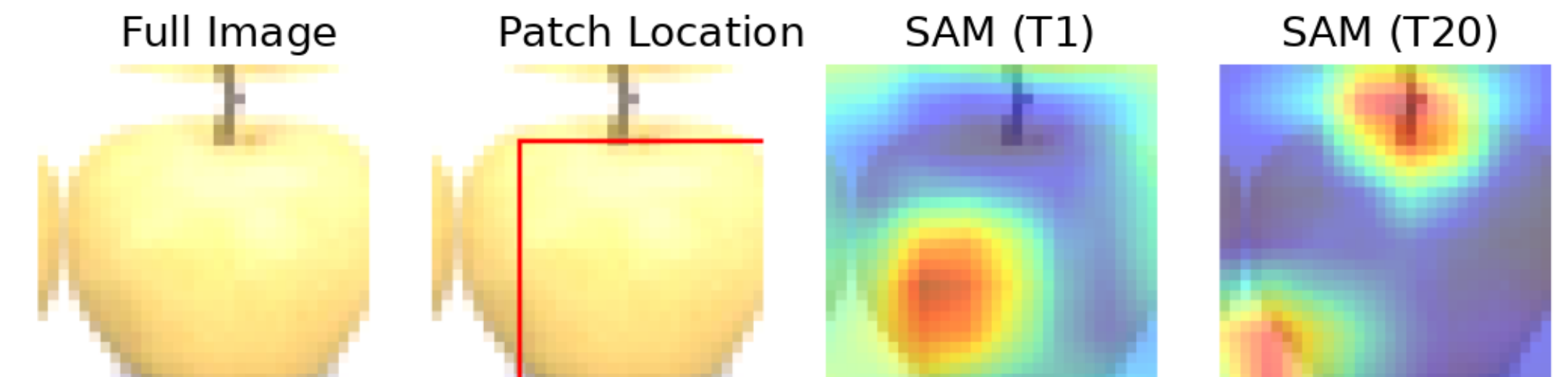}
         \caption{Task 1.}
         \label{fig:task1}
     \end{subfigure}
     \hfill
     \begin{subfigure}{0.48\linewidth}
         \centering
         \includegraphics[width=\linewidth]{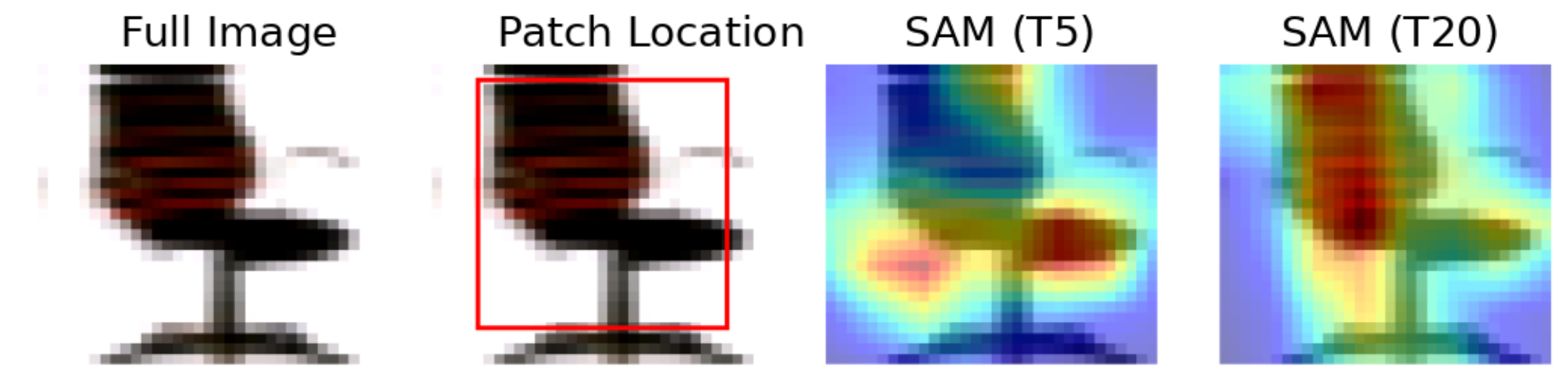}
         \caption{Task 5.}
         \label{fig:task5}
     \end{subfigure}
     \hfill
     \begin{subfigure}{0.48\linewidth}
         \centering
         \includegraphics[width=\linewidth]{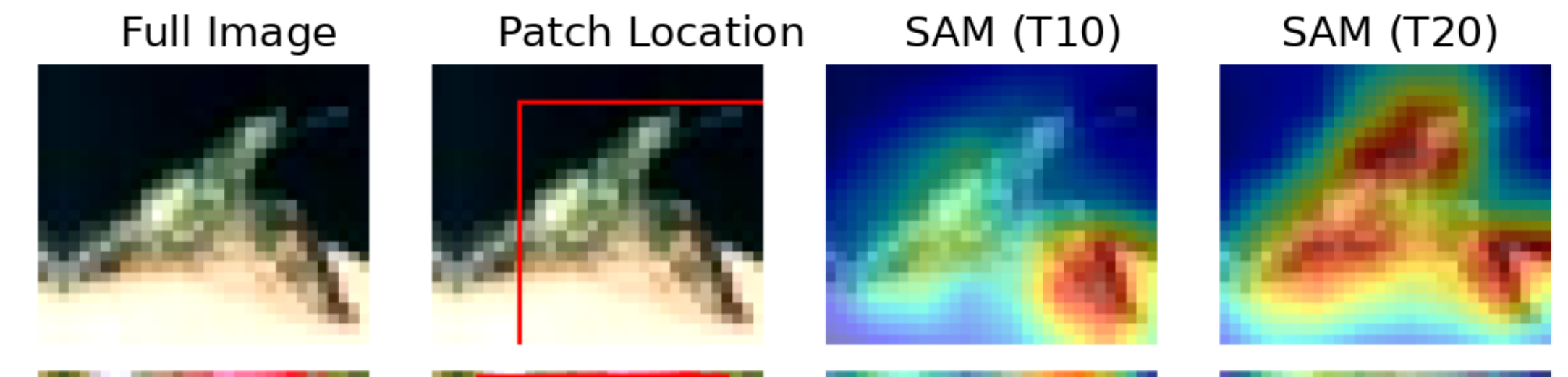}
         \caption{Task 10.}
         \label{fig:task10}
     \end{subfigure}
     \hfill
     \begin{subfigure}{0.48\linewidth}
         \centering
         \includegraphics[width=\linewidth]{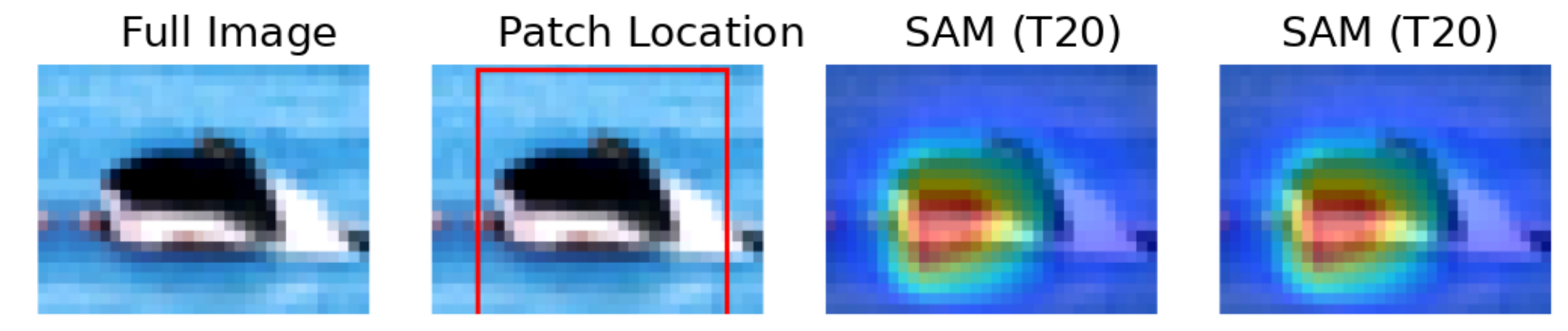}
         \caption{Task 20.}
         \label{fig:task20}
     \end{subfigure}
    \caption{Evolution of spatial attribution maps throughout continual learning. For Tasks 1, 5, 10, and 20, we show the input image, the selected patch location, the SAM generated immediately after learning the task from which the patch is selected, and the SAM obtained after training on all 20 tasks.}
    \label{fig:sams}
\end{figure}

Figure~\ref{fig:sams} compares SAMs generated immediately after learning a task with those obtained after completion of the task sequence. Across all tasks, the salient regions remain largely consistent, indicating preservation of task-relevant visual representations. Interestingly, for earlier tasks, the final SAMs are often more concentrated on the target object and less affected by background regions. This suggests that continual learning not only preserves relevant evidence but can also refine shared feature representations as the network is exposed to additional visual concepts.

\section{Conclusions}
We presented SNN-adapted EPR, a SAM-guided replay strategy for CL in deep SNNs. Experiments on Split-CIFAR10 and Split-CIFAR100 showed favorable accuracy–memory trade-offs compared to standard replay, particularly under tight memory budgets. Qualitative analysis further indicated that task-relevant representations are preserved and may become more localized over time. Limitations include dependence on SAM quality, the use of a fixed patch size, and evaluation on small-scale image classification benchmarks. Future work could explore pretrained backbones to improve early explanation quality, enabling better patch selection and memory initialization. Another direction is to make SAM more class-sensitive through output-conditioned scores or temporal attention, enabling the introduction of regularization terms to explicitly prevent explanation drift.

\begin{credits}
\subsubsection{\ackname} 
This work has been carried out while Michela Proietti and Federico Di Valerio were enrolled in the Italian National Doctorate on Artificial Intelligence and the PhD in Engineering in Computer Science, respectively, run by Sapienza University of Rome.

\subsubsection{\discintname}
The authors have no competing interests to declare that are relevant to the
content of this article.
\end{credits}

\bibliography{bibliography}
\end{document}